\documentclass[sigconf]{acmart}

\setcopyright{rightsretained}

\acmConference[]{SIGIR PatentSemTech Workshop}{July 15, 2022}{Madrid, Spain}
\copyrightyear{2022}

\begin{document}

\title{Patents Phrase to Phrase Semantic Matching Dataset}

\author{Grigor Aslanyan}
\email{aslanyan@google.com}
\affiliation{%
  \institution{Google Inc.}
}

\author{Ian Wetherbee}
\email{wetherbeei@google.com}
\affiliation{%
  \institution{Google Inc.}
}

\begin{abstract}
There are many general purpose benchmark datasets for Semantic Textual Similarity but none of them are focused on technical concepts found in patents and scientific publications. This work aims to fill this gap by presenting a new human rated contextual phrase to phrase matching dataset. The entire dataset contains close to $50,000$ rated phrase pairs, each with a CPC (Cooperative Patent Classification) class as a context. This paper describes the dataset and some baseline models.
\end{abstract}

\maketitle

\section{Introduction and Related Work}

Semantic Textual Similarity (STS) measures how similar two pieces of text are. STS is one of the most important tasks in Natural Language Processing (NLP) and there has been a significant amount of research in recent years in this domain. Benchmark datasets play the important role of allowing to consistently and fairly measure model improvements. There are multiple benchmark datasets for STS that are commonly used to measure the performance of state of the art models. Some notable examples are STS-B \cite{cer-etal-2017-semeval}, SICK \cite{marelli-etal-2014-sick}, MRPC \cite{dolan-brockett-2005-automatically}, and PIT \cite{xu-etal-2015-semeval}. However, these datasets are fairly general purpuse, and to the best of our knowledge there are currently no datasets that are focused on technical concepts found in patents and scientific publications. The somewhat related BioASQ challenge contains a biomedical question answering task \cite{283}.

This paper introduces a new human rated contextual phrase to phrase matching dataset focused on technical terms from patents. In addition to similarity scores that are typically included in other benchmark datasets we include granular rating classes similar to WordNet \cite{10.1145/219717.219748}, such as synonym, antonym, hypernym, hyponym, holonym, meronym, domain related.

The dataset was generated with focus on the following:
\begin{itemize}
\item Phrase disambiguation: certain keywords and phrases can have multiple different meanings. For example, the phrase "mouse" may refer to an animal or a computer input device. We have included a context CPC class that can help disambiguate the anchor and target phrase.
\item Adversarial keyword match: there are phrases that have matching keywords but are otherwise unrelated (e.g. “container section” $\rightarrow$ “kitchen container”, “offset table” $\rightarrow$ “table fan”). Many models will not do well on such data (e.g. bag of words models). Our dataset is designed to include many such examples.
\item Hard negatives: We created our dataset with the aim to improve upon current state of the art language models. Specifically, we have used the BERT model \cite{devlin-etal-2019-bert} to generate target phrases. So our dataset contains many human rated examples of phrase pairs that BERT may identify as very similar but in fact they may not be.
\end{itemize}

The full dataset is publicly available through Kaggle\footnote{ \href{https://www.kaggle.com/datasets/google/google-patent-phrase-similarity-dataset}{https://www.kaggle.com/datasets/google/google-patent-phrase-similarity-dataset}}. This dataset was used in the \emph{U.S. Patent Phrase to Phrase Matching} Kaggle competition\footnote{\href{https://www.kaggle.com/c/us-patent-phrase-to-phrase-matching}{https://www.kaggle.com/c/us-patent-phrase-to-phrase-matching}} from March 21 - June 20, 2022.


\section{Dataset Description}\label{sec:description}

\begin{table}
\begin{tabular}{ |c|c|c|c|c| } 
 \hline
 anchor & target & context & rating & score \\
 \hline
 acid absorption & absorption of acid & B08 & exact & 1.00 \\
 acid absorption & acid immersion & B08 & synonym & 0.75 \\
 acid absorption & chemically soaked & B08 & domain & 0.25 \\
 acid absorption & acid reflux & B08 & not rel. & 0.00 \\
 gasoline blend & petrol blend & C10 & synonym & 0.75 \\
 gasoline blend & fuel blend & C10 & hypernym & 0.50 \\
 gasoline blend & fruit blend & C10 & not rel. & 0.00 \\
 faucet assembly & water tap & A22 & hyponym & 0.50 \\
 faucet assembly & water supply & A22 & holonym & 0.25 \\
 faucet assembly & school assembly & A22 & not rel. & 0.00 \\
 \hline
\end{tabular}
\caption{A small sample of the data.}\label{table:sample}
\vspace{-7mm}
\end{table}

Each entry of the dataset contains two phrases - anchor and target, a context CPC class, a rating class, and a similarity score. A small sample of the dataset is shown in Table~\ref{table:sample}.

The entire dataset contains $48,548$ entries with $973$ unique anchors, split into a training ($75\%$), validation ($5\%$), and test ($20\%$) sets. When splitting the data all of the entries with the same anchor are kept together in the same set. There are $106$ different context CPC classes and all of them are represented in the training set.

We have used the following steps for generating the data. For each patent in the corpus we first extract important (salient) phrases. These are typically noun phrases (e.g. “fastener”, “lifting assembly”) or functional phrases (e.g. “food processing”, “ink printing”). Next, we keep only phrases that appear in at least 100 patents. We randomly sample around 1,000 phrases from the remaining phrases which become our anchor phrases. For each anchor phrase we find all of the matching patents and all of the CPC classes for those patents. From all of the matching CPC classes we randomly sample up to four. These become the context CPC classes for that anchor phrase. The target phrases come from two sources - pre-generated and rater generated.

We use two different methods for pre-generating target phrases - partial matching and a masked language model (MLM). For partial match we randomly select phrases from the entire corpus that partially match with the anchor phrase. This means that one or more of the tokens matches, but the whole phrase is different (e.g. “abatement” $\rightarrow$ “noise abatement”, “material formation” $\rightarrow$ “formation material”). For MLM we select sentences from the patents that contain a given anchor phrase, mask it out, and use a BERT model \cite{devlin-etal-2019-bert} to predict candidates for the masked portion of the text. All of the phrases are cleaned up before sending to the raters. This includes lowercasing and removal of punctuation and certain stopwords (e.g. "and", "or", "said").

The raters were asked to determine the similarity level between the two phrases given the context CPC class. They choose between five different levels of similarity - very high, high, medium, low, and not related. Each similarity level is further divided into different subclasses, such as hyponym (broad-narrow), hypernym (narrow-broad), antonym, domain related. The detailed description of the similarity levels and subclasses will be included with the public release of the data.

All of the pre-generated target phrases were independently rated by two raters. After completing the ratings they met and went over all of the non-matching ratings to discuss and agree on a final rating. Each rater separately generated new target phrases and gave ratings to them. We have merged all of the rater generated phrases together. We have left out the rare cases where the two raters generated the same target phrase with different ratings.

\section{Baselines}\label{sec:baselines}

\begin{table}
\begin{tabular}{ |c|c|c|c| } 
 \hline
 Model & Dim. & Pearson cor. & Spearman cor. \\
 \hline
 GloVe & 300 & 0.429 & 0.444 \\
 FastText & 300 & 0.402 & 0.467 \\
 Word2Vec & 250 & 0.437 & 0.483 \\
 BERT & 1024 & 0.418 & 0.409 \\
 Patent-BERT & 1024 & 0.528 & 0.535 \\
 Sentence-BERT & 768 & 0.598 & 0.577 \\
 \hline
\end{tabular}
\caption{Baseline model metrics.}\label{table:baselines}
\vspace{-7mm}
\end{table}

Table \ref{table:baselines} describe the performance of some common off the shelf models on the test data. We have only included dual-tower model architectures that perform an embedding of the anchor and target phrases separately and compute similarity using cosine distance. All of the models use mean pooling of individual keyword embeddings to get the full phrase embedding.

For GloVe \cite{pennington-etal-2014-glove} we have used the \emph{Wikipedia 2014 + Gigaword 5} model\footnote{\href{https://nlp.stanford.edu/projects/glove/}{https://nlp.stanford.edu/projects/glove/}}, for FastText \cite{joulin-etal-2017-bag} the \emph{wiki-news-300d-1M} model\footnote{\href{https://fasttext.cc/docs/en/english-vectors.html}{https://fasttext.cc/docs/en/english-vectors.html}} \cite{mikolov2018advances}, and for Word2Vec \cite{2013arXiv1301.3781M} the \emph{Wiki-words-250} model from TensorFlow Hub\footnote{\href{https://tfhub.dev/google/Wiki-words-250/2}{https://tfhub.dev/google/Wiki-words-250/2}}. For BERT \cite{devlin-etal-2019-bert} we have used the BERT-Large model from TensorFlow Hub\footnote{\href{https://tfhub.dev/tensorflow/bert_en_uncased_L-24_H-1024_A-16/4}{https://tfhub.dev/tensorflow/bert\_en\_uncased\_L-24\_H-1024\_A-16/4}}. For comparison, we have also included the publicly available BERT model pre-trained on patent data \cite{patent-bert} of the same size as BERT-Large. Finally, for Sentence-BERT \cite{reimers-2019-sentence-bert} we have used the \emph{all-mpnet-base-v2} pretrained model\footnote{\href{https://www.sbert.net/docs/pretrained_models.html}{https://www.sbert.net/docs/pretrained\_models.html}}.

The bag-of-words models do not perform very well, which is expected given the dataset structure (e.g. many matching terms with different meanings). The Patent-BERT model significantly outperforms the regular BERT model, which implies that generic pretrained models are not optimal for technical terms found in patents. However, we get the best results from the Sentence-BERT model. This is not entirely surprising since Sentence-BERT has been specifically fine tuned for the dual-tower architecture we are using for similarity.

\bibliographystyle{ACM-Reference-Format}
\bibliography{bibliography}


\begin{thebibliography}{13}


\ifx \showCODEN    \undefined \def \showCODEN     #1{\unskip}     \fi
\ifx \showDOI      \undefined \def \showDOI       #1{#1}\fi
\ifx \showISBNx    \undefined \def \showISBNx     #1{\unskip}     \fi
\ifx \showISBNxiii \undefined \def \showISBNxiii  #1{\unskip}     \fi
\ifx \showISSN     \undefined \def \showISSN      #1{\unskip}     \fi
\ifx \showLCCN     \undefined \def \showLCCN      #1{\unskip}     \fi
\ifx \shownote     \undefined \def \shownote      #1{#1}          \fi
\ifx \showarticletitle \undefined \def \showarticletitle #1{#1}   \fi
\ifx \showURL      \undefined \def \showURL       {\relax}        \fi
\providecommand\bibfield[2]{#2}
\providecommand\bibinfo[2]{#2}
\providecommand\natexlab[1]{#1}
\providecommand\showeprint[2][]{arXiv:#2}

\bibitem[\protect\citeauthoryear{Cer, Diab, Agirre, Lopez-Gazpio, and
  Specia}{Cer et~al\mbox{.}}{2017}]%
        {cer-etal-2017-semeval}
\bibfield{author}{\bibinfo{person}{Daniel Cer}, \bibinfo{person}{Mona Diab},
  \bibinfo{person}{Eneko Agirre}, \bibinfo{person}{I{\~n}igo Lopez-Gazpio},
  {and} \bibinfo{person}{Lucia Specia}.} \bibinfo{year}{2017}\natexlab{}.
\newblock \showarticletitle{{S}em{E}val-2017 Task 1: Semantic Textual
  Similarity Multilingual and Crosslingual Focused Evaluation}. In
  \bibinfo{booktitle}{\emph{Proceedings of the 11th International Workshop on
  Semantic Evaluation ({S}em{E}val-2017)}}. \bibinfo{publisher}{Association for
  Computational Linguistics}, \bibinfo{address}{Vancouver, Canada},
  \bibinfo{pages}{1--14}.
\newblock
\urldef\tempurl%
\url{https://doi.org/10.18653/v1/S17-2001}
\showDOI{\tempurl}


\bibitem[\protect\citeauthoryear{Devlin, Chang, Lee, and Toutanova}{Devlin
  et~al\mbox{.}}{2019}]%
        {devlin-etal-2019-bert}
\bibfield{author}{\bibinfo{person}{Jacob Devlin}, \bibinfo{person}{Ming-Wei
  Chang}, \bibinfo{person}{Kenton Lee}, {and} \bibinfo{person}{Kristina
  Toutanova}.} \bibinfo{year}{2019}\natexlab{}.
\newblock \showarticletitle{{BERT}: Pre-training of Deep Bidirectional
  Transformers for Language Understanding}. In
  \bibinfo{booktitle}{\emph{Proceedings of the 2019 Conference of the North
  {A}merican Chapter of the Association for Computational Linguistics: Human
  Language Technologies, Volume 1 (Long and Short Papers)}}.
  \bibinfo{publisher}{Association for Computational Linguistics},
  \bibinfo{address}{Minneapolis, Minnesota}, \bibinfo{pages}{4171--4186}.
\newblock
\urldef\tempurl%
\url{https://doi.org/10.18653/v1/N19-1423}
\showDOI{\tempurl}


\bibitem[\protect\citeauthoryear{Dolan and Brockett}{Dolan and
  Brockett}{2005}]%
        {dolan-brockett-2005-automatically}
\bibfield{author}{\bibinfo{person}{William~B. Dolan} {and}
  \bibinfo{person}{Chris Brockett}.} \bibinfo{year}{2005}\natexlab{}.
\newblock \showarticletitle{Automatically Constructing a Corpus of Sentential
  Paraphrases}. In \bibinfo{booktitle}{\emph{Proceedings of the Third
  International Workshop on Paraphrasing ({IWP}2005)}}.
\newblock
\urldef\tempurl%
\url{https://aclanthology.org/I05-5002}
\showURL{%
\tempurl}


\bibitem[\protect\citeauthoryear{Joulin, Grave, Bojanowski, and Mikolov}{Joulin
  et~al\mbox{.}}{2017}]%
        {joulin-etal-2017-bag}
\bibfield{author}{\bibinfo{person}{Armand Joulin}, \bibinfo{person}{Edouard
  Grave}, \bibinfo{person}{Piotr Bojanowski}, {and} \bibinfo{person}{Tomas
  Mikolov}.} \bibinfo{year}{2017}\natexlab{}.
\newblock \showarticletitle{Bag of Tricks for Efficient Text Classification}.
  In \bibinfo{booktitle}{\emph{Proceedings of the 15th Conference of the
  {E}uropean Chapter of the Association for Computational Linguistics: Volume
  2, Short Papers}}. \bibinfo{publisher}{Association for Computational
  Linguistics}, \bibinfo{address}{Valencia, Spain}, \bibinfo{pages}{427--431}.
\newblock
\urldef\tempurl%
\url{https://aclanthology.org/E17-2068}
\showURL{%
\tempurl}


\bibitem[\protect\citeauthoryear{Marelli, Menini, Baroni, Bentivogli, Bernardi,
  and Zamparelli}{Marelli et~al\mbox{.}}{2014}]%
        {marelli-etal-2014-sick}
\bibfield{author}{\bibinfo{person}{Marco Marelli}, \bibinfo{person}{Stefano
  Menini}, \bibinfo{person}{Marco Baroni}, \bibinfo{person}{Luisa Bentivogli},
  \bibinfo{person}{Raffaella Bernardi}, {and} \bibinfo{person}{Roberto
  Zamparelli}.} \bibinfo{year}{2014}\natexlab{}.
\newblock \showarticletitle{A {SICK} cure for the evaluation of compositional
  distributional semantic models}. In \bibinfo{booktitle}{\emph{Proceedings of
  the Ninth International Conference on Language Resources and Evaluation
  ({LREC}'14)}}. \bibinfo{publisher}{European Language Resources Association
  (ELRA)}, \bibinfo{address}{Reykjavik, Iceland}, \bibinfo{pages}{216--223}.
\newblock
\urldef\tempurl%
\url{http://www.lrec-conf.org/proceedings/lrec2014/pdf/363_Paper.pdf}
\showURL{%
\tempurl}


\bibitem[\protect\citeauthoryear{{Mikolov}, {Chen}, {Corrado}, and
  {Dean}}{{Mikolov} et~al\mbox{.}}{2013}]%
        {2013arXiv1301.3781M}
\bibfield{author}{\bibinfo{person}{Tomas {Mikolov}}, \bibinfo{person}{Kai
  {Chen}}, \bibinfo{person}{Greg {Corrado}}, {and} \bibinfo{person}{Jeffrey
  {Dean}}.} \bibinfo{year}{2013}\natexlab{}.
\newblock \showarticletitle{{Efficient Estimation of Word Representations in
  Vector Space}}.
\newblock \bibinfo{journal}{\emph{arXiv e-prints}}, Article
  \bibinfo{articleno}{arXiv:1301.3781} (\bibinfo{date}{Jan.}
  \bibinfo{year}{2013}), \bibinfo{numpages}{arXiv:1301.3781}~pages.
\newblock
\showeprint[arxiv]{cs.CL/1301.3781}


\bibitem[\protect\citeauthoryear{Mikolov, Grave, Bojanowski, Puhrsch, and
  Joulin}{Mikolov et~al\mbox{.}}{2018}]%
        {mikolov2018advances}
\bibfield{author}{\bibinfo{person}{Tomas Mikolov}, \bibinfo{person}{Edouard
  Grave}, \bibinfo{person}{Piotr Bojanowski}, \bibinfo{person}{Christian
  Puhrsch}, {and} \bibinfo{person}{Armand Joulin}.}
  \bibinfo{year}{2018}\natexlab{}.
\newblock \showarticletitle{Advances in Pre-Training Distributed Word
  Representations}. In \bibinfo{booktitle}{\emph{Proceedings of the
  International Conference on Language Resources and Evaluation (LREC 2018)}}.
\newblock


\bibitem[\protect\citeauthoryear{Miller}{Miller}{1995}]%
        {10.1145/219717.219748}
\bibfield{author}{\bibinfo{person}{George~A. Miller}.}
  \bibinfo{year}{1995}\natexlab{}.
\newblock \showarticletitle{WordNet: A Lexical Database for English}.
\newblock \bibinfo{journal}{\emph{Commun. ACM}} \bibinfo{volume}{38},
  \bibinfo{number}{11} (\bibinfo{date}{nov} \bibinfo{year}{1995}),
  \bibinfo{pages}{39–41}.
\newblock
\showISSN{0001-0782}
\urldef\tempurl%
\url{https://doi.org/10.1145/219717.219748}
\showDOI{\tempurl}


\bibitem[\protect\citeauthoryear{Pennington, Socher, and Manning}{Pennington
  et~al\mbox{.}}{2014}]%
        {pennington-etal-2014-glove}
\bibfield{author}{\bibinfo{person}{Jeffrey Pennington},
  \bibinfo{person}{Richard Socher}, {and} \bibinfo{person}{Christopher
  Manning}.} \bibinfo{year}{2014}\natexlab{}.
\newblock \showarticletitle{{G}lo{V}e: Global Vectors for Word Representation}.
  In \bibinfo{booktitle}{\emph{Proceedings of the 2014 Conference on Empirical
  Methods in Natural Language Processing ({EMNLP})}}.
  \bibinfo{publisher}{Association for Computational Linguistics},
  \bibinfo{address}{Doha, Qatar}, \bibinfo{pages}{1532--1543}.
\newblock
\urldef\tempurl%
\url{https://doi.org/10.3115/v1/D14-1162}
\showDOI{\tempurl}


\bibitem[\protect\citeauthoryear{Reimers and Gurevych}{Reimers and
  Gurevych}{2019}]%
        {reimers-2019-sentence-bert}
\bibfield{author}{\bibinfo{person}{Nils Reimers} {and} \bibinfo{person}{Iryna
  Gurevych}.} \bibinfo{year}{2019}\natexlab{}.
\newblock \showarticletitle{Sentence-BERT: Sentence Embeddings using Siamese
  BERT-Networks}. In \bibinfo{booktitle}{\emph{Proceedings of the 2019
  Conference on Empirical Methods in Natural Language Processing}}.
  \bibinfo{publisher}{Association for Computational Linguistics}.
\newblock
\urldef\tempurl%
\url{http://arxiv.org/abs/1908.10084}
\showURL{%
\tempurl}


\bibitem[\protect\citeauthoryear{{Srebrovic} and {Yonamine}}{{Srebrovic} and
  {Yonamine}}{[n.d.]}]%
        {patent-bert}
\bibfield{author}{\bibinfo{person}{Rob {Srebrovic}} {and} \bibinfo{person}{Jay
  {Yonamine}}.} \bibinfo{year}{[n.d.]}\natexlab{}.
\newblock \showarticletitle{{Leveraging the BERT algorithm for Patents with
  TensorFlow and BigQuery}}.
\newblock  (\bibinfo{year}{[n.\,d.]}).
\newblock
\urldef\tempurl%
\url{https://services.google.com/fh/files/blogs/bert_for_patents_white_paper.pdf}
\showURL{%
\tempurl}


\bibitem[\protect\citeauthoryear{Tsatsaronis, Balikas, Malakasiotis, Partalas,
  Zschunke, Alvers, Weissenborn, Krithara, Petridis, Polychronopoulos,
  Almirantis, Pavlopoulos, Baskiotis, Gallinari, Artieres, Ngonga, Heino,
  Gaussier, Barrio-Alvers, Schroeder, Androutsopoulos, and
  Paliouras}{Tsatsaronis et~al\mbox{.}}{2015}]%
        {283}
\bibfield{author}{\bibinfo{person}{George Tsatsaronis},
  \bibinfo{person}{Georgios Balikas}, \bibinfo{person}{Prodromos Malakasiotis},
  \bibinfo{person}{Ioannis Partalas}, \bibinfo{person}{Matthias Zschunke},
  \bibinfo{person}{Michael~R Alvers}, \bibinfo{person}{Dirk Weissenborn},
  \bibinfo{person}{Anastasia Krithara}, \bibinfo{person}{Sergios Petridis},
  \bibinfo{person}{Dimitris Polychronopoulos}, \bibinfo{person}{Yannis
  Almirantis}, \bibinfo{person}{John Pavlopoulos}, \bibinfo{person}{Nicolas
  Baskiotis}, \bibinfo{person}{Patrick Gallinari}, \bibinfo{person}{Thierry
  Artieres}, \bibinfo{person}{Axel Ngonga}, \bibinfo{person}{Norman Heino},
  \bibinfo{person}{Eric Gaussier}, \bibinfo{person}{Liliana Barrio-Alvers},
  \bibinfo{person}{Michael Schroeder}, \bibinfo{person}{Ion Androutsopoulos},
  {and} \bibinfo{person}{Georgios Paliouras}.} \bibinfo{year}{2015}\natexlab{}.
\newblock \showarticletitle{An overview of the BIOASQ large-scale biomedical
  semantic indexing and question answering competition}.
\newblock \bibinfo{journal}{\emph{BMC Bioinformatics}}  \bibinfo{volume}{16}
  (\bibinfo{year}{2015}), \bibinfo{pages}{138}.
\newblock
\urldef\tempurl%
\url{https://doi.org/10.1186/s12859-015-0564-6}
\showDOI{\tempurl}


\bibitem[\protect\citeauthoryear{Xu, Callison-Burch, and Dolan}{Xu
  et~al\mbox{.}}{2015}]%
        {xu-etal-2015-semeval}
\bibfield{author}{\bibinfo{person}{Wei Xu}, \bibinfo{person}{Chris
  Callison-Burch}, {and} \bibinfo{person}{Bill Dolan}.}
  \bibinfo{year}{2015}\natexlab{}.
\newblock \showarticletitle{{S}em{E}val-2015 Task 1: Paraphrase and Semantic
  Similarity in {T}witter ({PIT})}. In \bibinfo{booktitle}{\emph{Proceedings of
  the 9th International Workshop on Semantic Evaluation ({S}em{E}val 2015)}}.
  \bibinfo{publisher}{Association for Computational Linguistics},
  \bibinfo{address}{Denver, Colorado}, \bibinfo{pages}{1--11}.
\newblock
\urldef\tempurl%
\url{https://doi.org/10.18653/v1/S15-2001}
\showDOI{\tempurl}


\end{thebibliography}

\end{document}